\documentclass[conference]{IEEEtran}
\IEEEoverridecommandlockouts
\usepackage{cite}
\usepackage{amsmath,amssymb,amsfonts}
\usepackage{graphicx}
\usepackage{textcomp}
\usepackage{xcolor}
\usepackage{subfig}

\usepackage[noend]{algorithm2e}
\RestyleAlgo{ruled}
\usepackage{multirow}
\usepackage{enumitem}

\def\BibTeX{{\rm B\kern-.05em{\sc i\kern-.025em b}\kern-.08em
    T\kern-.1667em\lower.7ex\hbox{E}\kern-.125emX}}
\begin{document}

\title{Where did you tweet from? Inferring the origin locations of tweets based on contextual information}

\author{\IEEEauthorblockN{Rabindra Lamsal}
\IEEEauthorblockA{\textit{The University of Melbourne}\\
Melbourne, Australia \\
rlamsal@student.unimelb.edu.au}
\and
\IEEEauthorblockN{Aaron Harwood\textsuperscript{\textsection}}
\IEEEauthorblockA{\textit{The University of Melbourne}\\
Melbourne, Australia \\
arnhrwd@gmail.com}
\and
\IEEEauthorblockN{Maria Rodriguez Read}
\IEEEauthorblockA{\textit{The University of Melbourne}\\
Melbourne, Australia \\
maria.read@unimelb.edu.au}
}

\IEEEoverridecommandlockouts
\IEEEpubid{\makebox[\columnwidth]{978-1-6654-8045-1/22/\$31.00~\copyright2022 IEEE \hfill} \hspace{\columnsep}\makebox[\columnwidth]{ }}

\maketitle
\IEEEpubidadjcol

\begingroup\renewcommand\thefootnote{\textsection}
\footnotetext{Work done while the author was at the University of Melbourne.}
\endgroup

\begin{abstract}
Public conversations on Twitter comprise many pertinent topics including disasters, protests, politics, propaganda, sports, climate change, epidemics/pandemic outbreaks, etc., that can have both regional and global aspects. Spatial discourse analysis rely on geographical data. However, today less than 1\% of tweets are geotagged; in both cases---point location or bounding place information. A major issue with tweets is that Twitter users can be at location \textit{A} and exchange conversations specific to location \textit{B}, which we call the \textit{Location A/B} problem. The problem is considered solved if location entities can be classified as either origin locations (Location \textit{A}s) or non-origin locations (Location \textit{B}s). In this work, we propose a simple yet effective framework---the \textit{True Origin Model}---to address the problem that uses machine-level natural language understanding to identify tweets that conceivably contain their origin location information. The model achieves promising accuracy at country (80\%), state (67\%), city (58\%), county (56\%) and district (64\%) levels with support from a Location Extraction Model as basic as the \textit{CoNLL-2003}-based \textit{RoBERTa}. We employ a tweet contexualizer (\textit{locBERT}) which is one of the core components of the proposed model, to investigate multiple tweets' distributions for understanding Twitter users' tweeting behavior in terms of mentioning origin and non-origin locations. We also highlight a major concern with the currently regarded gold standard test set (ground truth) methodology, introduce a new data set, and identify further research avenues for advancing the area.

\end{abstract}


\section{Introduction}

Microblogging platforms such as Twitter and Sina Weibo have a significant impact on public discourse; these platforms have raised the possibility of people participating in public conversations, and have become an active source of information during both day-to-day life and in unusual circumstances ~\cite{lamsal2022socially} such as earthquakes, floods, hurricanes, tsunamis, cyclones, wildfires, and pandemics. Especially during emergency events, the number of conversations generated on these platforms reaches hundreds of thousands and even millions in long-term events such as the ongoing COVID-19 pandemic. For instance, during an ongoing disastrous event, people tend to use these platforms excessively, as they share their safety status and exchange conversations to query the safety status of their friends and family. People also share what they have seen, felt, or heard from others. During such critical hours of a disaster, the use of social media peaks at unprecedented levels, and based on these public conversations, first responders and decision-makers can visualize a more comprehensive real-time picture of the situation to aid in formulating actionable plans. Collection, processing, and analysis of such an enormous amount of socially generated data is not practical to manually undertake and warrants the use of automated stream processing systems for a timely understanding of how an event is unfolding---\textit{situation awareness}---concerning its temporal and spatial dimension~\cite{lamsal2022socially}.

\begin{figure}[!h]
    \centering
    \includegraphics[width=0.48\textwidth]{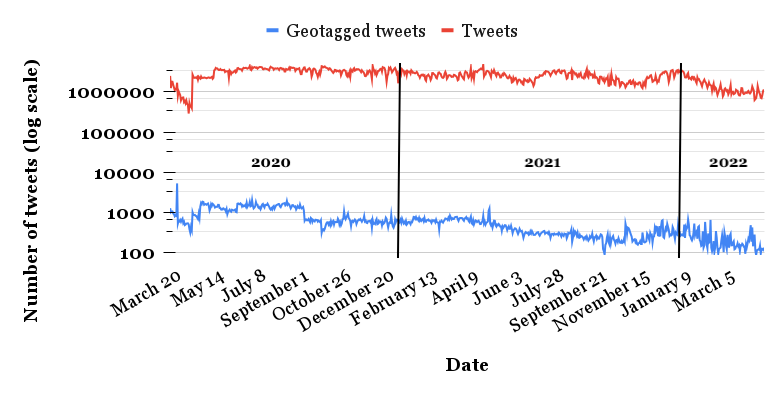}
    \caption{Daily distribution of tweets in the \textit{COV19Tweets} \cite{781w-ef42-20} and \textit{GeoCOV19Tweets} \cite{fpsb-jz61-20} Datasets. Today less than 1\% of tweets are
geotagged.}
    \label{tweet-counts}
\end{figure}

Spatial analysis rely on geographic location data. Location-involved analysis concerning microblogging conversations include discourse-based forecasting, movement and travel pattern analysis, behavior-based interests and intent analysis, generation of task and region-specific heatmaps, environment monitoring, transportation planning, urban planning, mobility analysis, crisis management and so on~\cite{lamsal2022socially}. However, today less than 1\% of tweets are geotagged~\cite{lamsal2021design, qazi2020geocov19}, be it with point coordinates or bounding place information. For instance, the billion-scale COVID-19 tweets dataset at IEEE~\cite{781w-ef42-20}, curated since the inception of the pandemic, reports 480k tweets being geotagged with point coordinates~\cite{fpsb-jz61-20} out of 2 billion tweets collected (the daily distribution of tweets for both the datasets is shown in Figure \ref{tweet-counts}). Therefore, it is critical to extract the \emph{origin locations} of as many tweets to decrease the possibility of bias during location-involved discourse analysis such as \cite{lamsal2022twitter}.

\subsection{The geolocation extraction problem}

Numerous tweet objects in the \textit{tweet data dictionary}\footnote{https://developer.twitter.com/en/docs/twitter-api/v1/data-dictionary/} assist in the extraction of location information. If a tweet is geotagged, it either contains the \textit{coordinates} object to represent the tweet's point location or the \textit{place} object to represent a place. The place information available in the \textit{place} object might not be the origin location of the tweet. For non-geotagged tweets, tweet objects such as \textit{text}, \textit{user:location}, \textit{user:description}, \textit{entities:media}, and \textit{lang} (for regional languages) are helpful for location inference. Country or region-specific contexts in \textit{profile\_image\_url\_https} and \textit{profile\_banner\_url} objects also can assist in location inference. The \textit{utc\_offset} and \textit{time\_zone} objects have been used in the past for inferring probable locations; however, these objects have become deprecated in recent versions of Twitter's APIs.

\textbf{Location A/B problem.}
The major issue with determining the origin locations of tweets is that a Twitter user can be at location \textit{A} and participate in a public discourse specific to location \textit{B}---which we call the \emph{Location A/B} problem \cite{lamsalabproblem}. Tweets comprise two types of geolocations, i.e., origin and non-origin locations. Origin locations are the ones from where the tweet has been made; a tweet such as ``I love the weather here in Melbourne" suggests that the user was at the mentioned geolocation at the time the tweet was made. However, tweets such as ``Stop this war. \#russia" and ``Melbourne is a beautiful city." do not provide sufficient evidence to conclude that the mentioned geolocations are the origin locations. Also, tweets with point coordinates might not necessarily originate from the respective point sources, which eventually contributes to the \textit{Location A/B} problem. We discuss this issue later in Section \ref{location-evidence} with supporting data.

The textual data contains the \emph{primary} context for origin and non-origin locations, therefore, this study investigates if machine-level natural language understanding is effective in identifying tweets that possibly contain origin locations. We rely exclusively on the \textit{text} object, since the other location-inferable tweet objects, discussed earlier, enclose location identities that might be independent to any particular tweet.

The paper is organized as follows: Related work is discussed in Section~\ref{related-work}. Section~\ref{approach} discusses the fundamentals---geocoding, locations vectors and data. Section~\ref{experimentation} proposes a framework for addressing the \textit{Location A/B} problem. Section~\ref{eval-framework} evaluates the proposed framework. Section~\ref{tweets-analysis} investigates Twitter users’ behavior in terms of
mentioning origin and non-origin locations in tweets, and Section~\ref{conclusion} concludes the paper.

\section{Related Work}
\label{related-work}
Categorizing locations as origin or non-origin involves the extraction of location entities as an initial step. There have been numerous studies that aimed at inferring location entities from tweets. The majority of the studies involve using named-entity recognizers \cite{gelernter2011geo, lingad2013location} for extracting toponyms from the \textit{text} tweet object. Additional tweet objects, such as \textit{user:location}, \textit{user:description}, \textit{user:url}, \textit{user:timezone}, have also been incorporated as possible spatial indicators~\cite{schulz2013multi}.

In~\cite{li2014effective}, Li et al. proposed a location identification method that combines multiple tweets of a user to identify top-$k$ locations of the user and further refines the identified locations to come up with top-$k$ locations for a tweet. In~\cite{li2014fine}, Li and Sun extracted fine-grained locations from tweets based on mentioned point-of-interest (POI) while exploiting the Foursquare check-ins done by users from the same geolocation. Similarly, Mahmud et al. \cite{mahmud2012tweet} proposed an ensemble approach that involved extracting features such as words, hashtags, and places, leveraging domain knowledge, and combining statistical and heuristics classifications to infer the home locations of Twitter users.

Davis Jr et~al.~\cite{davis2011inferring} inferred locations of tweets based on analyzing the follow/follower associations amongst Twitter users; the idea involved obtaining reciprocal relationships by taking an intersection between the followers of a user and the people the user follows. In~\cite{priedhorsky2014inferring}, Priedhorsky et al. proposed an inference method based on Gaussian mixture models (GMMs) trained on geotagged tweets. For inferring the origin location of a new tweet, the previously trained GMMs (for each unique $n$-gram, a 2D GMM was fitted to model it geographic distribution) were combined for the $n$-grams present in the tweet. In~\cite{yuan2013and},  Yuan et al. proposed a probabilistic model that incorporates the who (user), where (geo), when (time), and what (terms) contexts for inferring the location of a user given a tweet and a time. Zubiaga et~al.~\cite{zubiaga2017towards} created eight different classifiers based on \textit{user:location}, \textit{user:lang}, \textit{user:timezone}, \textit{lang}, \textit{user:utc\_offset}, \textit{user:name}, \textit{user:description}, and \textit{text} tweet objects to infer the country of origin of tweets. Similarly, Lau et~al.~\cite{lau2017end} proposed an end-to-end neural network that uses \textit{text}, \textit{created\_at}, \textit{user:utc\_offset}, \textit{user:timezone}, \textit{user:location}, \textit{user:created\_at} tweet objects to predict the geolocation of a tweet.

Content similarity has also been helpful in location inference problems. In~\cite{kinsella2011m},  Kinsella et al. estimated distributions of terms for each location considered in the study and assigned a new tweet to the location with the highest probability by sampling from the distributions. In~\cite{paraskevopoulos2015fine}, Paraskevopoulos and Palpanas used similarity measure between a tweet and a set of geotagged tweets alongside their temporal characteristics to infer its origin at city level. Similarly, Li et~al.~\cite{li2018location} split users' timelines into different clusters, each cluster implying a distinct location at city level. Next, they classified each cluster into its predefined class, such that future tweets are assigned city-level locations accordingly. In~\cite{ikawa2012location}, Ikawa et~al. proposed an association-based approach to identifying locations of tweets based on the relations between locations and their relevant keywords from past tweets.

Gazetteer-based searches for n-grams have been employed extensively to identify location candidates. In~\cite{middleton2013real}, Middleton et~al. performed location token matching for geoparsing tweet content. Similarly, in~\cite{malmasi2015location}, Malmasi and Dras proposed Noun Phrase extraction and n-gram-based matching to detect mentions of locations in tweets.  Similarly, in~\cite{qazi2020geocov19}, Qazi et~al. performed n-gram-based matching to identify location candidates. The authors extended their work~\cite{imran2022tbcov} with the use of language-based NER models. The choice of gazetteer remains to be a trade-off between recall and efficiency; Dutt et~al.~\cite{dutt2018savitr} evaluated the performance of the most favored gazetteers, i.e., \textit{GeoNames} and \textit{OpenStreetMaps}.

\subsection*{Contributions of this study}

\begin{itemize}[leftmargin=*]
    \item As per our extensive literature search, this study is the first to address the \textit{Location A/B} problem; we introduce a tweet-processing framework---the \textit{True Origin Model}---for the task. The framework has two components, Location Extraction Model (\textit{LEM}) and \textit{locBERT}. The location inference methods/models in the existing literature are the \textit{LEM} candidates. The \textit{LEM} candidates do not consider \textit{Location A/B} context and tweets such as “Stop this war. \#russia” also get geotagged and included, which introduces biases in spatial analysis or discourse-based models. To address this issue, the next component in the framework, \textit{locBERT}, enriches geotagging by extracting presence/absence of origin location information.
    \item We introduce the concept of \textit{tweets origin location evidence dataset} for training language models that aim to identify the origin locations of tweets.
    \item We investigate multiple distributions of tweets to explore Twitter users’ tweeting behavior in terms of mentioning origin and non-origin locations
    \item We identify and present significant research avenues for future work advancing the area.
\end{itemize}

\section{The Fundamentals}
\label{approach}
\subsection{Forward/reverse geocoding}
\label{apiserver}
\textit{Forward Geocoding}, or simply \textit{Geocoding}, refers to the process of converting addresses (such as ``700 Swanston Street, Carlton, Melbourne, VIC'') to their geographic coordinates (such as longitude 144.96449828 and latitude -37.80011159). The addresses may not necessarily be structured as the aforementioned example; for instance, ``melbourne uni'' converts to longitude 144.96130134 and latitude -37.7970796, and ``chapman avenue glenroy'' converts to longitude 144.9121865 and latitude -37.7064932. \textit{Reverse geocoding}, on the other hand, is the process of converting longitude/latitude pairs into human-readable addresses.

\textbf{Planet-level API endpoints.}
\label{planet-api}
There are services that provide forward and reverse geocoding endpoints---Google's Geocoding API, Mapbox, positionstack, and Nominatim, to name a few. The existing services either limit the number of search hits to their endpoints or become expensive to pay for when experimenting with hundreds of thousands of queries. Therefore, we built planet-level forward (\textit{search endpoint}) and reverse (\textit{reverse endpoint}) geocoding endpoints for this study on a Linux machine with 24 VCPUs and 216 gigabytes memory. The endpoints are powered by OpenStreetMap data and elastic search\footnote{https://en.wikipedia.org/wiki/Elasticsearch}, with a search index of 61 gigabytes\footnote{https://download1.graphhopper.com/public/} (when compressed). Both the endpoints return a payload in JSON format with the following keys: coordinates, properties:\{country, city, countrycode, postcode, type, street, district, name, state\}.

\subsection{Location vectors}
\label{section-locationvectors}
This study aims to extract geographic locations from tweets at five different levels---district, county, city, state, and country. Given a location, $LOC$, a location vector, $V_{LOC}$, is returned by the forward geocoding endpoint, with values for each component of $V_{LOC}$. For instance, if $LOC$ is a district, its location vector, $V_{LOC}$, will have valid entries for each component; however, if $LOC$ is a state, the granular components---city, county, district---will have \textit{NULL} entries, and state and country components will have valid entries. 
However, during reverse geocoding, for a set of longitude/latitude pairs provided to the reverse geocoding endpoint, all the components of $V_{LOC}$ are returned with valid entries.
%

\subsection{Twitter Data}
\begin{figure}
    \centering
    \includegraphics[width=0.475\textwidth]{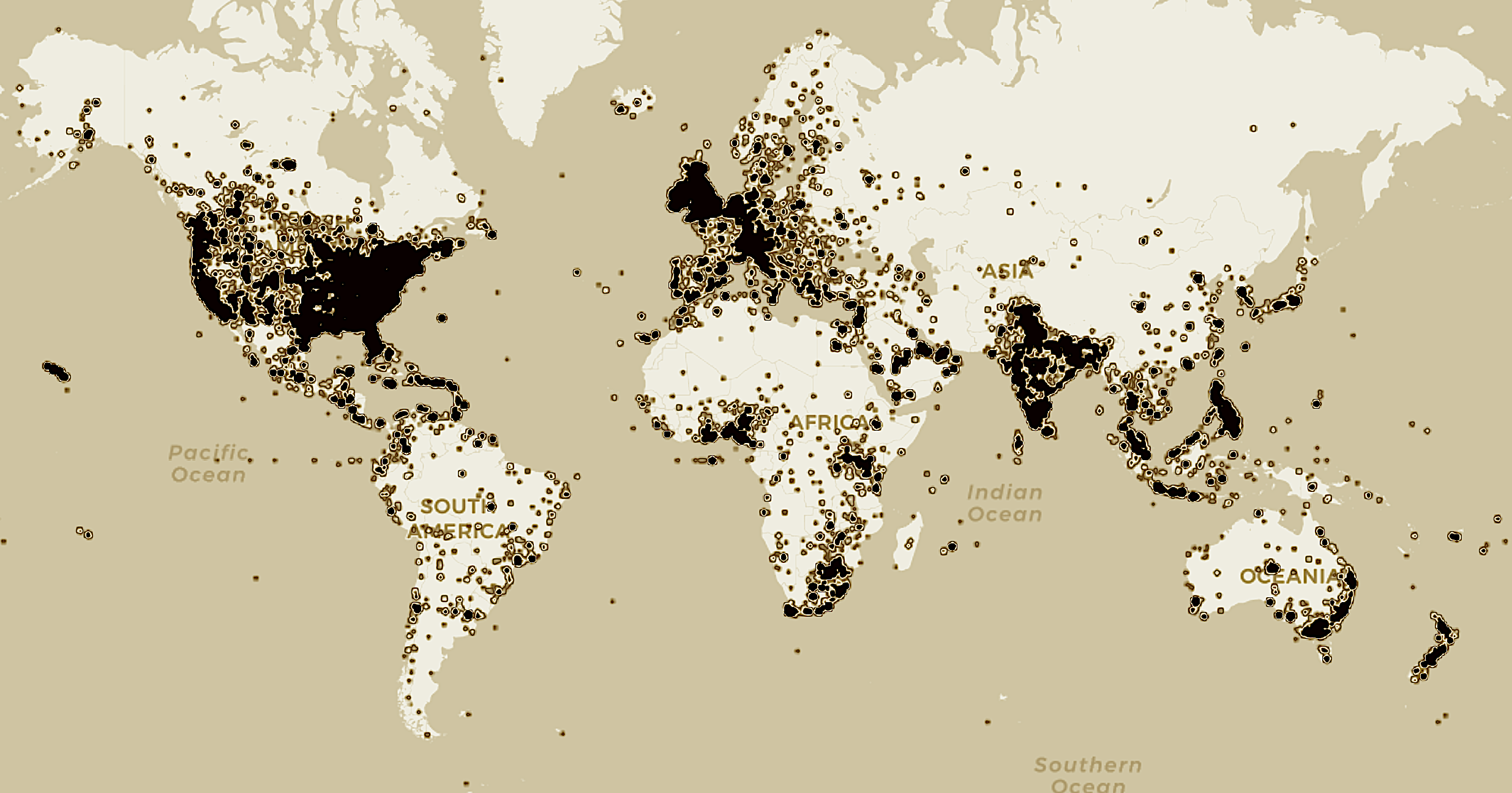}
    \caption{Geographical coverage\protect\footnotemark~of the \textit{GeoCOV19Tweets Dataset}~\cite{fpsb-jz61-20}.}
    \label{coverage}
\end{figure}

The primary source of data for this study is the \textit{GeoCOV19Tweets Dataset}~\cite{fpsb-jz61-20}, which is the geo-version of the billion-scale \textit{COV19Tweets Dataset}~\cite{781w-ef42-20} that is currently being maintained and updated daily at IEEE DataPort and is one of the longest-running tweet collections regarding the COVID-19 pandemic. The dataset is being curated while tracking Twitter's near-real-time feed for coronavirus-related tweets using more than 90 keywords and hashtags that are closely related to the pandemic; experimenting on this dataset helps capture the overall conversational dynamics of the pandemic such that the proposed method can be implemented on a similar scale or smaller COVID-19 specific datasets. The dataset~\cite{fpsb-jz61-20} contains tweet identifiers of geo-tagged (point-location) COVID-19 specific tweets; this study makes use of the dataset's latest version, i.e., the tweets collected over the period March 20, 2020~--~April 28, 2022. Twitter's data re-distribution policy restricts the sharing of data except for tweet identifiers, user identifiers, and message identifiers; the identifiers need to be hydrated to re-create a raw tweet corpus locally~\cite{lamsal2021design}. The tweet identifiers in the \textit{GeoCOV19Tweets Dataset} were hydrated using Twitter's \textit{Tweets lookup}\footnote{https://developer.twitter.com/en/docs/twitter-api} endpoint. Figure \ref{coverage} shows the geographical coverage of the dataset.

\textbf{Issue with gold standard test set methodology.}
\label{issue-gold-set}
The currently regarded gold standard test set methodology for evaluating the performance of tweet origin location identification models uses sets of geotagged tweets with point locations that are extracted using the \textit{coordinates:coordinates} tweet object. However, during this study, we observed that around 71\% of tweets with point locations on Twitter are in fact Instagram posts re-shared by their respective authors. This is based on inspection of the \textit{source} tweet object on one of the longest-running (25+ months) COVID-19-specific tweets repositories~\cite{fpsb-jz61-20, 781w-ef42-20}. Instagram aims primarily at sharing photos and videos, with less of an emphasis on real-time posts compared to Twitter's feed which has a focus on ``as-they-happen'' personal and news updates. When the tweets in~\cite{fpsb-jz61-20} were scrutinized, the majority of tweets shared from Instagram were identified as stories/updates from the past (for example---\textit{\#throwback to a visit from our hiring partner! \#repost \#thankyou \#repost walgreensjobs. Pre-Covid, one of our pharmacy teams visited the Miami campus of Florida Vocational Institute to speak to future pharmacy…}). Furthermore, third-party platforms that provide auto-share and scheduling features also give rise to the \textit{Location A/B} problem. Tweets created through the native application can also exhibit \textit{Location A/B} issues. These observations suggest that even the geotagged tweets (with point locations) might not necessarily originate from the respective point sources; concluding that the contexts of tweets should be critically considered when using the geotagged tweets as ground truth for any particular study.

\footnotetext{https://live.rlamsal.com.np/}

\section{The Proposed Framework and Experiments}
\label{experimentation}

Our proposed framework---the \textit{True Origin Model}---is valid for all possible cases where a set of tweet identifiers are available, primarily: (i) pre-collected sets of tweet identifiers, (ii) searching or streaming of raw twitter data. The first case requires the use of the Tweet lookup endpoint, while the second case requires requesting search or streaming endpoints with valid queries (combination of keywords, hashtags, filters/conditions).

\begin{figure*}
    \centering
    \includegraphics[width=1\textwidth]{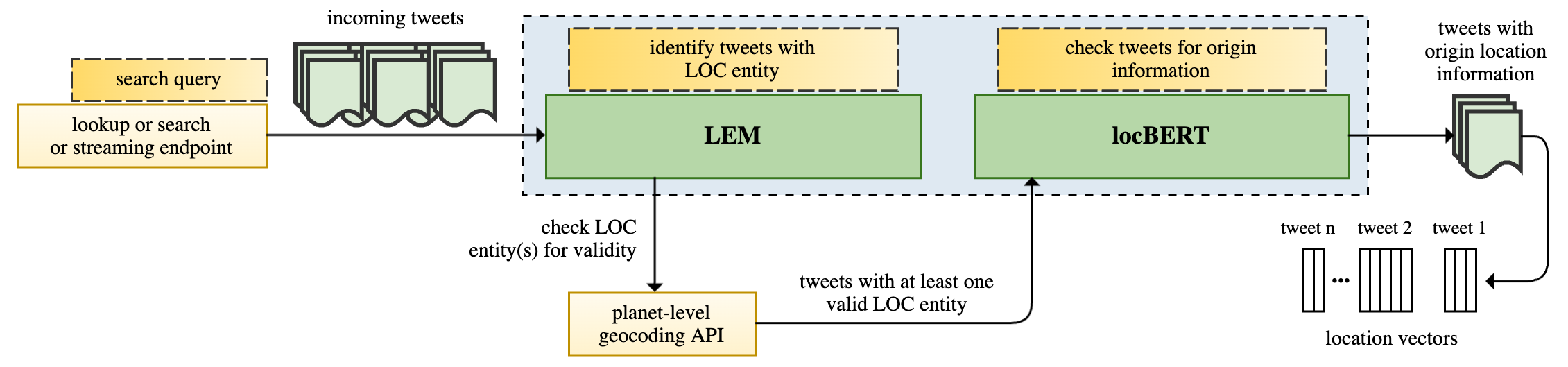}
    \caption{The \textit{True Origin Model}.}
    \label{TOM}
\end{figure*}

Figure \ref{TOM} illustrates the overall view of the proposed \textit{True Origin Model}. The hydrated tweets or search/stream-returned tweets are provided as inputs to the core of the model that consists of three major components: (i) a Location Extraction Model (\textit{LEM}), (ii) two planet-level geocoding endpoints, and (iii) \textit{locBERT} (a tweet
contextualizer). The \textit{LEM} component, discussed in more detail in Section~\ref{lem}, extracts location mentions present within the \textit{text} tweet object. The input tweets are checked for location mentions, and only the tweets with at least one location entity are considered for the next phase. Each identified location mention is validated by a planet-level forward geocoding endpoint (discussed in Section~\ref{planet-api}). If the geocoding endpoint returns a valid response for an identified location mention, we consider the location as valid. Tweets with at least one valid location mention are provided as input to \textit{locBERT} (discussed in detail in Section~\ref{locbert})---a transformer-based~\cite{vaswani2017attention} model that we propose for identifying tweets that conceivably contain their origin location information. Tweets identified as containing origin information are processed further to extract location vectors based on the identified valid location mentions. Finally, we take the majority of similar vector components for each tweet to extract the probable origin location. 

Next, we discuss \textit{LEM} and \textit{locBERT} in detail.

\subsection{Location Extraction Model}
\label{lem}

The Location Extraction Model (\textit{LEM}), in general, is a location-entity extraction model that identifies location mentions in unstructured texts or documents. For instance, from the tweet ``\textit{Love this bunch. Off to the Park for a morning walk in nature \#NJ \#NewJersey. First COVID Times visit to \#Manhattan. HRH @MENTION \& I, are in \#NewYork for a few days.... \#NYC here we are...}'', a \textit{LEM} candidate identifies the following tokens as location entities: \textit{Park}, \textit{NJ}, \textit{NewJersey}, \textit{Manhattan}, \textit{NewYork}, \textit{NYC}.

The location inference methods/models that we discussed in the related work section are \textit{LEM} candidates. For experimentation purposes, and to avoid reinventing the wheel, we forked some of the most popular transformer-based named-entity extraction models from Hugging Face \cite{wolf2019huggingface}. Note that, the \textit{LEM} is an unrestricted component that can incorporate any geotagger for extracting location entities. We randomly sampled 60k tweets from the \textit{GeoCOV19Tweets Dataset} using pandas' \textit{DataFrame.sample}\footnote{https://pandas.pydata.org/} procedure (for reproducibility: random\_state=19). We experimented with the four most forked transformer-based \cite{vaswani2017attention} LARGE architecture pre-trained models on Hugging Face, namely XLM-RoBERTa \cite{conneau2019unsupervised}, RoBERTa\cite{liu2019roberta}, BERT \cite{devlin2018bert}, ELECTRA \cite{clark2020electra}, on the sampled tweets corpus. These models were fine-tuned on the CoNLL-2003 dataset~\cite{sang2003introduction} to extract location entities. The performance of each model on the sampled tweets corpus is summarized in Table~\ref{perf-ner}. We consider only the LARGE architectures since they seem to outperform the base variants; using LARGE architectures justify our use case since we aim to identify as many location mentions in the \textit{text} tweet object. Algorithm~\ref{lem-pseudo} presents the pseudocode for performance comparison of the \textit{LEM} candidates.

\begin{algorithm}
\footnotesize
\DontPrintSemicolon

\SetKwFunction{FMain}{location\_validity}
\SetKwProg{Fn}{Function}{:}{}
\Fn{\FMain{$location$}}{
Send request to the forward geocoding endpoint.\;
\textbf{if} len(response['features']) == 0 \textbf{then} validity = False \;
\textbf{else} validity = True\;
\KwRet validity\;
}

\SetKwFunction{FMain}{locations\_extract}
\SetKwProg{Fn}{Function}{:}{}
\Fn{\FMain{tweet}}{

result = $model(tweet)$\;
Maintain a list \textit{loc\_details}.\;
Maintain a list \textit{locations}.\;
\For{each identified location in \textit{result}}{
    Add the identified location to \textit{locations} list\;
    Maintain a count of identified location(s).\;
    \If{location\_validity(identified location) == True}{
        Maintain a count of valid identified location(s).
    }
}
Append locations and maintained counts to loc\_details\;
\KwRet loc\_details
}
\;

\#\#Experimenting on the tweet corpus\#\#\;

dataframe.tweet\_column.apply(lambda x: $locations\_extract(x)$)

\caption{Pseudocode for performance comparison of \textit{LEM} candidates}
\label{lem-pseudo}
\end{algorithm}

The results in Table~\ref{perf-ner} show that RoBERTa outperforms all models in all four cases: identifying the most number of (i) location mentions, (ii) tweets with at least one location mention, (iii) valid locations, and (iv) tweets with at least one valid location. XLM-RoBERTa benchmarks second and is followed by BERT and ELECTRA in all four cases. However, there is another dimension to these performance scores---the most number of invalid outcomes. In Table~\ref{invalids}, we list the number of invalids identified as valids by each model. A location is considered invalid when the planet-level geocoding endpoint does not return a valid JSON response. And a tweet is invalid when the tweet does not contain any valid location(s). In the case of invalid outcomes, BERT performed the best with under 1k invalids and is followed by ELECTRA. However, for XLM-RoBERTa and RoBERTa, the numbers are significantly higher. We see similar outcomes for the invalid number of tweets, with ELECTRA benchmarking first. But, in this study, we are not concerned with identifying less number of invalids; instead, we aim to maximize the number of identified location mentions; therefore, we select RoBERTa as \textit{LEM} for our \textit{True Origin Model}.

\begin{table}[h!]
\centering
  \caption{Performance of the transformer-based location extraction models on COVID-19 English language tweets corpus. \textit{Notes}: $^a$number of locations identified in the tweet corpus, $^b$number of tweets identified with at least one location, $^c$number of valid locations amongst the identified ones, $^d$number of tweets identified with at least one valid location.}
  \label{perf-ner}
  \begin{tabular}{p{3.2cm}| p{0.9cm}|p{0.9cm}|p{0.8cm} |l}
    \hline
    \textbf{Model} & \textbf{\#idloc}$^a$ & \textbf{\#tloc}$^b$ & \textbf{\#vloc}$^c$ & \textbf{\#tvloc}$^d$ \\
    \hline
    XLM-RoBERTa & 75,609 & 37,964 & 71,980 & 37,410\\
    \hline
    RoBERTa & 78,454 & 38,027 & 73,537 & 37,492 \\
    \hline
    BERT & 72,474 & 37,045 & 71,601 & 36,863 \\
    \hline
    ELECTRA & 67,987 & 34,227 & 67,078 & 34,122 \\
  \hline
\end{tabular}
\end{table}

\begin{table}[h!]
\centering
  \caption{Invalids identified as valids by the transformer-based location extraction models. \textit{Notes}: $^a$location(s) for which the planet-level geocoding endpoint returned invalid response, $^b$tweet(s) without any valid location(s).}
  \label{invalids}
  \begin{tabular}{p{3.2cm}|p{2.2cm}|p{2.2cm}}
    \hline
   \textbf{Model}  & \textbf{invalid locations}$^a$ & \textbf{invalid tweets}$^b$\\
    \hline
    XLM-RoBERTa & 3,629 & 554\\
    \hline
    RoBERTa & 4,917 & 535\\
    \hline
    BERT & 873 & 182\\
    \hline
    ELECTRA & 909 & 105\\
  \hline
\end{tabular}
\end{table}

\subsection{locBERT}
\label{locbert}

\subsubsection{Origin location evidence}
\label{location-evidence}
On Twitter, besides the tweets geotagged with point location, there are a set of tweets that can be assigned origin location based on how they have been composed. For instance, consider the following tweets:

\textbf{Example 1}: \textit{Fun 3 Mile Interval Run along the Riverwalk!
Supporting our beautiful city of Chicago, our beloved Tamale Guy who has recently been hospitaliized, fighting Covid-19 and yes, the USPS, I will always need stamps.}

\textbf{Example 2}: \textit{@MENTION and I traveled to \#Miami at the beginning of June for his 30th Bday EMOJI and our 3 year anniversary EMOJI! Many of you were asking for travel tips amid COVID-19, so I finally put a blog together to help you…}

\textbf{Example 3}: \textit{Continuing to support local businesses, I drove to Berchman’s Brewing Company this evening which, sadly, is closing their downtown Yakima taproom at the end of July due to the pandemic EMOJI and I purchased a grunt…}


Among these examples, the first and third have origin location evidence, while the second does not. The problem of identifying origin location evidence necessitates natural language understanding; rule-based approaches are ineffective when the context has to be decided from informally written and grammatically incorrect texts with less than 280 characters. For instance, tweets containing the keyword ``today'' might be seen as including evidence for origin location; however, the context changes when the discussion is about something from the past, such as a ``throwback'', or is simply a news article. Besides, a major issue with conversations generated on Twitter is their source. When the tweets in the \textit{GeoCOV19Tweets Dataset} were checked for their original source, we noticed that out of 403.9k tweets, 71.10\% of them were shared on Twitter from Instagram by the authors of the respective Instagram posts. Instagram aims primarily at sharing media (photos/videos), while Twitter is used for instantaneous personal and news updates and promotes real-time posts. Instagram-post-turned-tweets might not necessarily originate from the locations referenced in the tweets. Furthermore, services with auto share and scheduling features, such as \textit{dlvrit} and \textit{Hootsuite}, also seem to generate a significant number of tweets on Twitter. We provide a list of the top 15 geotagged tweets’ sources in Table \ref{tweets-sources-list}. Therefore, even in the case of geotagged tweets, the point location provided in the \textit{tweet data dictionary} might not necessarily be the origin location. Taking these factors into consideration, we design a transformer-based language model to implement machine-level natural language understanding in the identification of tweets that conceivably contain their origin location information.

\begin{table}[h]
\centering
\caption{Top 15 geotagged tweets' sources.}
\label{tweets-sources-list}
\begin{tabular}{p{2.7cm}|l|p{2.7cm}|l}
\hline
\textbf{Source} & \textbf{\%} & \textbf{Source} & \textbf{\%}\\
\hline
Instagram & 71.1\% & Twitter for iPhone & 0.69\%\\
\hline
dlvr.it & 13.52\% & Untappd & 0.65\%\\
\hline
Hootsuite Inc. & 2.10\% & heapevents.info & 0.54\%\\
\hline
Tweetbot for iOS & 1.59\% & TweetCaster (iOS) & 0.37\%\\
\hline
Foursquare & 1.10\% & Tweetlogix & 0.32\%\\
\hline
TweetCaster (Android) & 1.04\% & Covid Genie & 0.28\%\\
\hline
Squarespace & 1.01\% & Foursquare Swarm & 0.27\%\\
\hline
FUNcation & 0.77\% & &\\
\hline
\end{tabular}
\end{table}

\subsubsection{Training data}
\label{annotated-data}

From the \textit{GeoCOV19Tweets Dataset}, we use two independent annotators to manually annotate 2,800 tweets (\textit{tweets origin location evidence dataset}) into two classes based on the presence and absence of origin location information: (i) true origin, (ii) low evidence, analogous to the examples presented in Section~\ref{location-evidence} and Appendix. A tweet was annotated as ``true origin'' only if it gives a clear context for the presence of origin location; otherwise, it was annotated as ``low evidence'', even if the context was ambiguous. The annotated dataset had 887 ``true origin'' tweets and 1913 ``low evidence'' tweets---the dataset is imbalanced in the ratio [1.58, 0.73]. The Cohen's kappa coefficient\footnote{sklearn.metrics.cohen\_kappa\_score} was computed for measuring inter-annotator agreement among the two independent annotators---the statistic was $0.83$, which can be interpreted as an almost-perfect agreement~\cite{mchugh2012interrater}.

\subsubsection{Training locBERT candidates}
We preprocess the annotated tweets as follows: (i) replace HTML entities with their character representation; such as ``\&amp;'' to ``\&'', (ii) replace URLs with HTTPURL token, (iii) replace user mentions with \@USER token, (iv) replace emoji\footnote{https://pypi.org/project/emoji/} with EMOJI token, (v) clean unnecessary spaces, indentations, paragraph breaks. Next, the dataset is partitioned into train (80\%), validation (10\%), and test (10\%) sets (for reproducibility: random\_state=19). As the candidate models for \textit{locBERT}, we fine-tuned the BASE architectures of BERT~\cite{devlin2018bert}, DistilBERT~\cite{sanh2019distilbert}, ELECTRA~\cite{clark2020electra}, RoBERTa~\cite{liu2019roberta}, XLM-RoBERTa~\cite{conneau2019unsupervised}, XLNet~\cite{yang2019xlnet}, and BERTweet~\cite{nguyen2020bertweet} on the annotated dataset using the Hugging Face library. The following configurations were considered during fine-tuning of the \textit{locBERT} candidates: Number of training epochs: $30$, Learning rate: $1e-6$, Training batch size: $8$, Maximum sequence length:  $70$, Optimizer: AdamW\footnote{https://pytorch.org/docs/stable/generated/torch.optim.AdamW.html}.

The results from the experiments are summarized in Table~\ref{results1}. The fine-tuned BERTweet outperformed all other models in terms of F${_1}$ for both classes. BERTweet is a language model pre-trained using RoBERTa pre-training procedure on a large-scale English language tweet corpus of 845 million tweets and an additional 23 million COVID-19 specific tweets. The results show that pre-training language models on domain-specific text corpora perform better than the models trained on ``standard'' text corpora. We use the fine-tuned BERTweet as \textit{locBERT} of our \textit{True Origin Model}. Refer to Appendix for a sample of tweets that were incorrectly classified by \textit{locBERT} during its evaluation. Understanding the context is not straightforward as in sentences such as  ``I love the weather here in Melbourne",  ``Stop this war. \#russia" and ``Melbourne is a beautiful city."---some real-world examples are provided in Appendix.

\begin{table}[h]
\centering
\caption{Performance of \textit{locBERT} candidates}
\label{results1}
\begin{tabular}{p{3.5cm}|p{2cm}|p{2cm}}
\hline
\textbf{Model}  & \textbf{F$_1$ true origin} & \textbf{F$_1$ low evidence}  \\
\hline
BERT  & 0.65 & 0.86\\
\hline
DistilBERT  & 0.61 &  0.82\\
\hline

ELECTRA  & 0.65 & 0.82\\
\hline

RoBERTa  & 0.70 & \textbf{0.87}\\
\hline

XLM-RoBERTa  & 0.66 & 0.85 \\
\hline

XLNet  & 0.70 & 0.86\\
\hline

BERTweet  & \textbf{0.74}  & \textbf{0.87}\\

\hline
\end{tabular}
\end{table}

\section{Evaluation of the proposed framework}
\label{eval-framework}
In this section, we evaluate \textit{LEM}, planet-level endpoints, and our proposed \textit{locBERT} together as a pipeline, shown in Figure~\ref{TOM}. The standard test data for this study is the \textit{GeoCOV19Tweets Dataset}. Since our method incorporates the context of tweets, it is, therefore, valid to use the tweets present in the \textit{GeoCOV19Tweets Dataset} for the evaluation.

\subsection{Location vectors}
Following the proposed workflow of the \textit{True Origin Model} in Figure~\ref{TOM}, the tweets that were identified by RoBERTa to have at least one valid location (in Section~\ref{lem}) were provided as inputs to \textit{locBERT} for categorizing the tweets as: (i) true origin, or (ii) low evidence (as discussed in Section~\ref{annotated-data}). However, there were a set of invalid locations identified by the RoBERTa model that were marked as valid by the geocoding endpoint. Therefore, as a further pre-processing step, we filtered locations that had: (i) character length less than 1,  (ii) only numeric digits, and (iii) solely a generic definition of a region, i.e., ``city'', ``earth'', ``europe'', ``asia'', ``americas'', ``africa'', ``world'', ``town'', ``county'', ``district''. After this step, 37.1k out of the 38k tweets were available for evaluation purposes. Those tweets were then input through \textit{locBERT}; out of the 37.1k tweets, \textit{locBERT} classified 14.9k tweets as ``true origin'' and the remaining 22k as ``low evidence''. There were instances where a longitude/latitude pair returned a \textit{NULL} location vector. Also, there were cases where the location(s) identified by \textit{LEM} returned a \textit{NULL} location vector. In both of these cases combined, 410 ``true origin'' class tweets were filtered out with 14,574 tweets, hereafter called the \textit{Eval Set}, available for evaluation. Table~\ref{eval-overview} gives the overview of the pre-processing, classification, and filtration steps discussed above.

\begin{table}[h!]
\centering
\caption{Overview of the tweets for evaluation. \textit{Notes}: $^a$tweets identified with at least one location by RoBERTa.}
\label{eval-overview}
\begin{tabular}{p{5.5cm}|l}
\hline
\textbf{Total tweets} & 60,000 \\
\hline
\textbf{Tweets with at least one valid location}$^a$ & 38,027\\
\hline
\textbf{Tweets after pre-processing} & 37,102\\
\hline
\multirow{2}{*}{\textbf{\textit{locBERT} categorization}}&true origin: 14,984\\
    \cline{2-2}
    &low evidence: 22,118\\
\hline
\textbf{Available tweets for evaluation} & 14,574\\
\hline
\end{tabular}
\end{table}

\subsection{Conclusive location vectors}
For each tweet in \textit{Eval Set}, we query the reverse geocoding endpoint with the longitude/latitude pair to compute its ground-truth location vector (a list for a single tweet). 
Similarly, for the location(s) identified in each tweet, we compute the location vector(s) (a list of lists for a single tweet) querying the forward geocoding endpoint. We then take the majority of similar vector components to compute a conclusive location vector. The conclusive location vector is compared with the ground-truth vector for evaluation. We provide an overview of the \textit{True Origin Model} as pseudocode in Algorithm~\ref{tom-pseudo}. The results from the evaluation are summarized in Table~\ref{eval}.

\begin{algorithm}
\footnotesize
\DontPrintSemicolon

$extract\_vector()$ \;
$extract\_vector\_reverse()$ \;
Maintain a dictionary country\_dict for country ISO code/name pairs.\;

\SetKwFunction{FMain}{identity\_location\_vector}
\SetKwProg{Fn}{Function}{:}{}
\Fn{\FMain{$locations$}}{
Maintain a list locations\_vector.\;

\For{each location in locations}{
Append $extract\_vector(location)$ to locations\_vector\;
}

Maintain lists district, county, city, state, country\;

\For{each vector in locations\_vector}{
Append vector[0] to the list district, vector[1] to the list county, vector[2] to the list city, vector[3] to the list state, vector[4] to the list country\;
}

\SetKwFunction{FMain}{vote}
\SetKwProg{Fn}{Function}{:}{}
\Fn{\FMain{list\_to\_count}}{
Maintain a frequency distribution of items in list\_to\_count.\;
}
\KwRet the most frequent item except None\;

Call $vote()$ for lists district, county, city, state, country\;

}
\KwRet most common item for each list as [district, county, city, state, country]\;

\SetKwFunction{FMain}{expand\_and\_check\_validity}
\SetKwProg{Fn}{Function}{:}{}
\Fn{\FMain{locations}}{

Maintain a list processed\_locations.

\For{each location in locations}{

Filter locations if they are generic names and $len(location)$$<$2.\;
Remove unwanted characters and symbols.\;
Append location to the list processed\_locations.\;
}
}
\KwRet the resulting list locations\;

\;
\#\#evaluating with geotagged tweets\#\#

Reverse geocode longitude/latitude pair to address vector using $extract\_vector\_reverse()$.\;

Preprocess locations using $expand\_and\_check\_validity()$ and normalize country ISO codes based on the dictionary country\_dict.\;

Generate location vectors of the locations extracted by \textit{LEM} using $identity\_location\_vector()$.\;

Compare results obtained from $extract\_vector\_reverse()$ and $identity\_location\_vector()$.\;

\caption{Pseudocode for the \textit{True Origin Model}}
\label{tom-pseudo}
\end{algorithm}

\begin{table}[h!]
\centering
\caption{Performance of the \textit{True Origin Model}}
\label{eval}
\begin{tabular}{p{2.1cm}|l|l|l|l|l}
\hline
& \textbf{Country} & \textbf{State} & \textbf{City} & \textbf{County} & \textbf{District}\\
\hline
\textbf{Correct} & 11,669 & 8,659 & 6,530 & 6,000 & 4229\\
\hline
\textbf{Incorrect} & 2,905 & 4,159 & 4733 & 4,585 & 2369\\
\hline
\textbf{Accuracy} & 80\% & 67.5\% & 58\% & 56.6\% & 64\%\\
\hline
\end{tabular}
\end{table}

\subsection{Comparison with existing studies}
It is not justifiable to compare the performance of \textit{locBERT} with the existing geotagging methods (which are the \textit{LEM}s) since \textit{locBERT}'s sole aim is to enrich the geotagging process by identifying the presence/absence of the origin location information. If we process $N$ tweets through existing geotaggers, we get $N$ geotagging results, with tweets such as ``Stop this war. \#russia" also getting geotagged. However, \textit{locBERT} ignores such tweets and geotags only $(O)*N$ number of tweets, where $O$ is the proportion of tweets with origin location evidence.

Also, previous studies use \textit{text} tweet object alongside objects such as \textit{created\_at}, \textit{user:location}, \textit{user:description}, \textit{user:url}, \textit{user:time\_zone}, \textit{user:lang}, \textit{user:utc\_offset}, and \textit{user:name}. The \textit{utc\_offset} and \textit{time\_zone} objects have been used extensively in the past as location indicators; however, these objects have become deprecated and return \textit{NULL} values in the tweet payload today. In their absence, it has become critical to explore the possibilities within the \textit{text} tweet object more than ever. The proposed model could have considered incorporating additional tweet objects for extracting more location vectors to increase its location prediction accuracy. However, the involvement of location indicators other than the \textit{text} tweet object creates forecast biases when tweets originate from places different than suggested by the indicators. The next-generation true origin models should focus on critically utilizing the \textit{text} tweet object; similar to \textit{user:timezone} and \textit{user:utc\_offset}, the currently available location indicators might become deprecated in the future, but raw content will continue to be available. Models trained solely on the \textit{text} object also have applications outside Twitterverse.

\section{Tweets analysis with \textit{locBERT}} 
\label{tweets-analysis}

\begin{figure*}[h!]

\setkeys{Gin}{width=0.24\linewidth}
\centering
\subfloat[Overall distribution ($N=403.9k$)]{\includegraphics{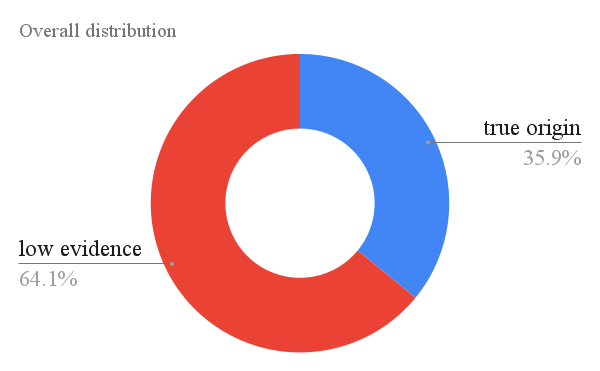}}
\subfloat[\textit{possibly\_sensitive} object]{\includegraphics{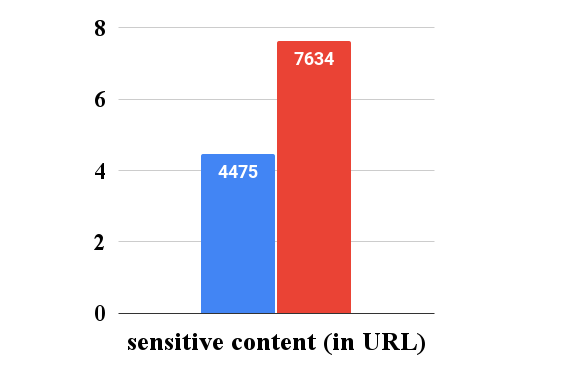}}
\subfloat[\textit{media} object]{\includegraphics{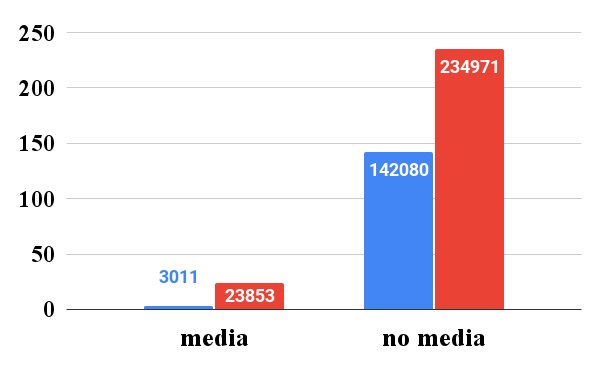}}
\subfloat[\textit{user\_verified} object]{\includegraphics{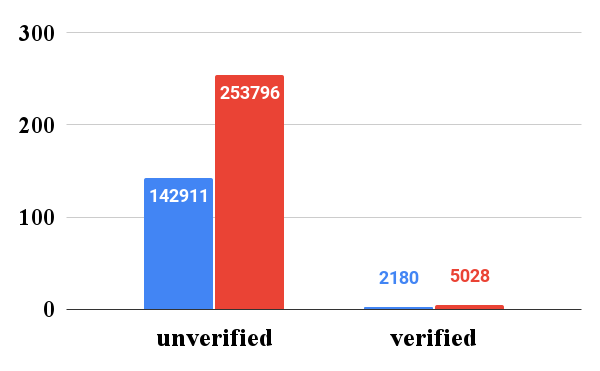}}

\setkeys{Gin}{width=0.49\linewidth}
\subfloat[\textit{user\_followers\_count} object]{\includegraphics{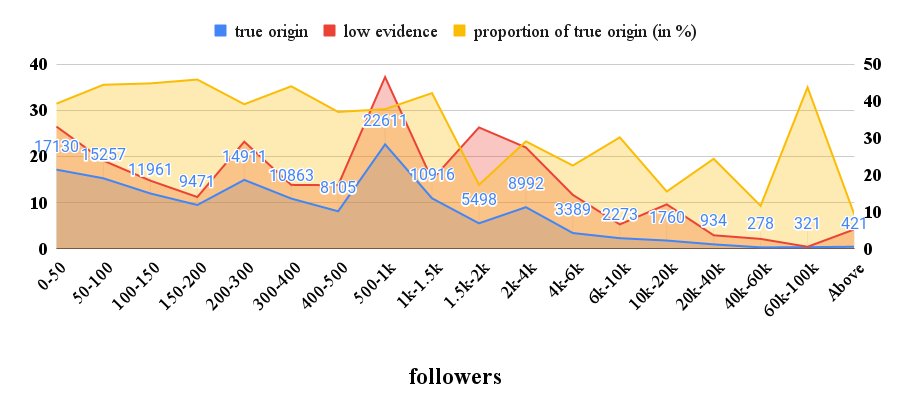}\label{followers-trend}}
\subfloat[\textit{user\_friends\_count} object]{\includegraphics{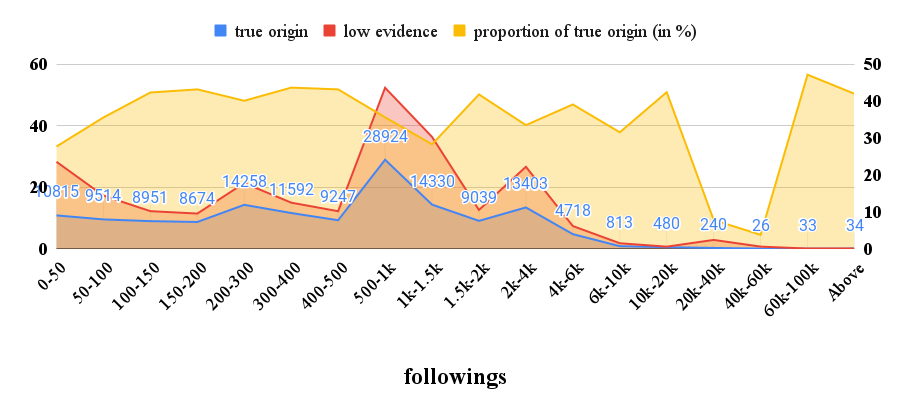}\label{followings-trend}}

\setkeys{Gin}{width=0.49\linewidth}
\subfloat[\textit{user\_created\_at} object]{\includegraphics{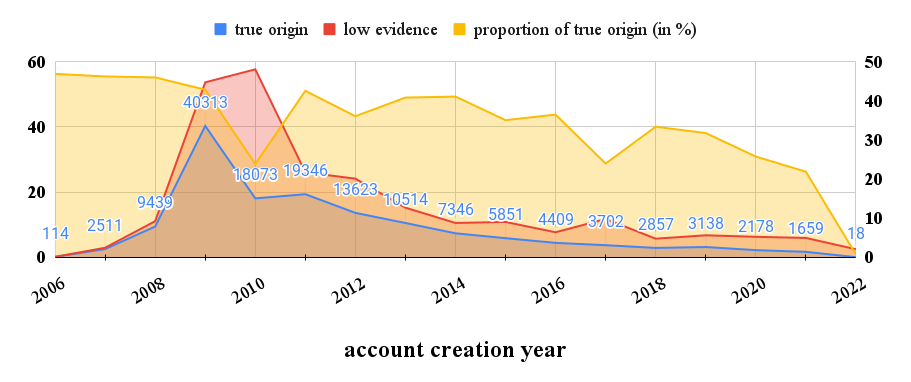}\label{joined-trend}}
\subfloat[\textit{country} object]{\includegraphics{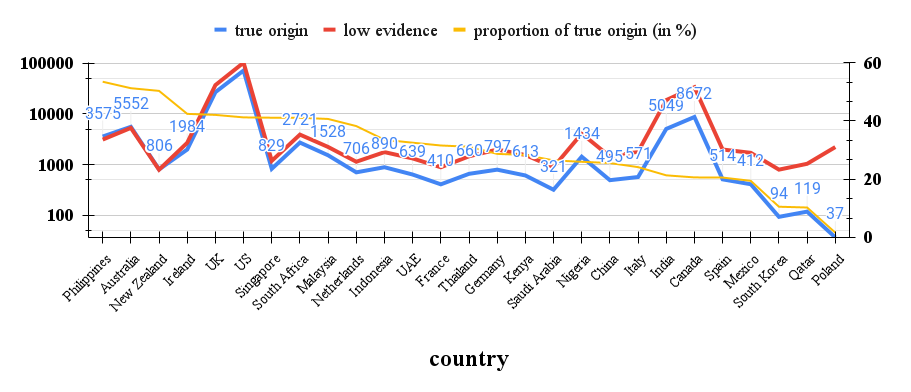}\label{geodetails}}

\caption{Distribution of tweets identified by \textit{locBERT} based on different tweets objects. For the Figures 4b-h, the vertical axis represents the \textit{frequency of tweets} and the scales are $1000x$ for Figures 4b-g and $log$ for Figure 4h. Figure \ref{geodetails} presents data for the most frequent 27 countries. The right y-axis in Figures 4e-g represent the proportion of true origin tweets.}
\label{all-dist-locbert}
\end{figure*}

We employed \textit{locBERT} on the \textit{GeoCOV19Tweets Dataset} ($N=403.9k$) to understand the distribution of Twitter users' behavior towards mentioning origin and non-origin locations, while tweeting, based on the following tweet objects: \textit{possibly\_sensitive}, \textit{media}, \textit{user\_verified}, \textit{user\_followers\_count}, \textit{user\_friends\_count}, \textit{user\_created\_at}, and \textit{country}. Figure~\ref{all-dist-locbert} presents multiple distributions from the experiments. We make numerous deductions based on these distributions. We report that 35.9\% of the geotagged tweets on Twitter mention origin locations, and in all the experimented cases, the number of tweets with origin locations is less than the number of tweets that mention non-origin locations. We observed that only 2.99\% of tweets contain sensitive content in shared URLs and the proportion of true origin and low evidence tweets is approximately equal to what has been reported in the overall distribution. We noticed that out of 403.9k tweets 6.65\% of them had media items (photos, videos, or GIFs), and amongst the media-attached tweets, 11.2\% were true origin tweets, and, in the case of non-media, 37.68\% of them were true origin tweets. However, in the case of verified and unverified Twitter accounts, we did not observe a significant difference in the proportion---unverified had 36\% and verified had 30\% tweets with origin locations.

Next, we created 18 bands to break down the number of \textit{followers} and \textit{followings} data. It is evident from Figure~\ref{followers-trend} that users with huge followers tend to tweet less, while users with 500-1k followers seem to be posting the highest number of tweets and are followed by users in the 0-50 and 200-300 bands. As the number of tweets decreases with the increase in the number of followers, the tweets with origin locations also seem to be following a similar trend. Also in terms of \textit{followings} (Figure \ref{followings-trend}), we observe a similar behavior. 

Further, we created a distribution for \textit{account creation year} versus the \textit{frequency of tweets}. Figure~\ref{joined-trend} shows that the users who joined Twitter between 2008--2014 have significant participation in the public discourse. However, the same timeline demonstrates a presence of a notable difference in the proportion of true origin and low evidence tweets. Users who joined Twitter between 2006--2008, and 2014--present seem to be the ones mentioning origin locations comparatively higher in terms of proportion.  Furthermore, we created a geolocation-based (country) distribution, and observed a relatively higher presence of low evidence tweets for almost all countries except Australia and the Philippines.

We investigated the influence of the tweets' sources on the generation of the true origin and low evidence tweets. Sorted by tweet frequency, Instagram tops the list with a presence of 37.61\% low evidence tweets (and 33.48\% true origin tweets), followed by \textit{dlvr.it}, \textit{Hootsuite Inc.}, \textit{Tweetbot for iOS}, \textit{TweetCaster for Android}, and \textit{Squarespace}, all generating low evidence tweets. After Instagram, \textit{Foursquare} and \textit{Untappd} are the second and third major sources of true origin tweets. These findings in terms of sources show that even the tweets geotagged with point coordinates cannot be considered ground truth data for evaluating models such as the one proposed in this study if the context for the presence and absence of origin location information is left out.

\section{Conclusion}
\label{conclusion}

In this paper, we introduced a tweet-processing framework---the \textit{True Origin Model}---for addressing the \textit{Location A/B} problem. The proposed approach involved a location extraction model (\textit{LEM}), two planet-level geocoding endpoints, and a tweets contextualizer working together in a pipeline. Overall, we: (\emph{i}) experimented with the four most forked transformer-based NER models as \textit{LEM} candidates; (\emph{ii}) created self-hosted instances of two planet-level geocoding endpoints; and (\emph{iii}) designed a transformer-based model (\textit{locBERT}) to identify tweets that possibly contain origin locations. \textit{locBERT} achieved promising results across country (80\%), state (67.5\%), city (58\%), county (56.6\%), and district (64\%) levels, with \textit{LEM} as basic as the \textit{CoNLL-2003}-based \textit{RoBERTa}. We briefly discussed the issue with the area's currently regarded gold standard test set methodology. We employed \textit{locBERT} to investigate multiple tweets' distributions for understanding Twitter users' tweeting behavior in terms of mentioning origin and non-origin locations. Finally, we presented some significant research avenues to take on.

\subsection*{Future directions}
There are multiple significant research avenues to take on. The best candidate for \textit{locBERT} achieved an average F$_1$ of 0.805 (F$_1$ on ``true origin'' class was 0.74) and this suggests there remains room for improvement. Further, the issue with common places needs to be addressed. For instance, the place name ``Victoria'' is a state in Australia as well as a city in British Columbia, Canada. Such commonality issues can be addressed (to an extent) by scrutinizing the location entities mentioned in chronologically earlier ``true origin'' tweets in the user's timeline. Twitter provides three timeline endpoints\footnote{https://developer.twitter.com/en/docs/twitter-api/tweets/timelines/} to fetch tweets from a specific Twitter account. Furthermore, there is a need for an effective processing of ubiquitous location entities such as \textit{Hall}, \textit{Ring Road}, \textit{Central park}, etc. We scrutinized a sample of incorrect identifications and observed that the presence of generic/ubiquitous location entities significantly decreases the model's performance since the location vector of each identified location entity contributes equally for generating the conclusive location vector. Table~\ref{locations-generic} presents a sample of location entities associated with incorrect identifications.
    
\begin{table}[!h]
\centering
  \caption{Some example generic/ubiquitous location entities associated with the incorrect identifications}
  \label{locations-generic}
  \begin{tabular}{p{8.4cm}}
    \hline
    Saint, PeelSt, Manship, arock, ari, The Bush, East Lynn, Sola, AVATAR, Kings Inn, The Dungeon, Willows, West End, ola, Brandon, Central Park, Steam Pub, Bay, blackrock, states, East Cobb, Soka, North City, Beaches, Ring Road, The Oak, hall, The Junction, Port, lakeview East, Ambassador Park, Poolside, Cafe, Home, International Airport, Gym, coast, paradise, South Bank\\
  \hline
\end{tabular}
\end{table}
    
\section*{Acknowledgement}
This study is supported by the Melbourne Research Scholarship from the University of Melbourne, Australia. This study was also supported by use of the Nectar Research Cloud, a collaborative Australian research platform supported by the NCRIS-funded Australian Research Data Commons (ARDC). Also, thanks to Digital Ocean\footnote{https://www.digitalocean.com} for providing infrastructures needed for the curation of the \textit{COV19Tweets Dataset}\footnote{https://ieee-dataport.org/open-access/coronavirus-covid-19-tweets-dataset}.

\bibliographystyle{IEEEtranS}
\bibliography{mybibfile}

\begin{thebibliography}{10}
\providecommand{\url}[1]{#1}
\csname url@samestyle\endcsname
\providecommand{\newblock}{\relax}
\providecommand{\bibinfo}[2]{#2}
\providecommand{\BIBentrySTDinterwordspacing}{\spaceskip=0pt\relax}
\providecommand{\BIBentryALTinterwordstretchfactor}{4}
\providecommand{\BIBentryALTinterwordspacing}{\spaceskip=\fontdimen2\font plus
\BIBentryALTinterwordstretchfactor\fontdimen3\font minus
  \fontdimen4\font\relax}
\providecommand{\BIBforeignlanguage}[2]{{%
\expandafter\ifx\csname l@#1\endcsname\relax
\typeout{** WARNING: IEEEtranS.bst: No hyphenation pattern has been}%
\typeout{** loaded for the language `#1'. Using the pattern for}%
\typeout{** the default language instead.}%
\else
\language=\csname l@#1\endcsname
\fi
#2}}
\providecommand{\BIBdecl}{\relax}
\BIBdecl

\bibitem{clark2020electra}
K.~Clark, M.-T. Luong, Q.~V. Le, and C.~D. Manning, ``Electra: Pre-training
  text encoders as discriminators rather than generators,'' \emph{arXiv
  preprint arXiv:2003.10555}, 2020.

\bibitem{conneau2019unsupervised}
A.~Conneau, K.~Khandelwal, N.~Goyal, V.~Chaudhary, G.~Wenzek, F.~Guzm{\'a}n,
  E.~Grave, M.~Ott, L.~Zettlemoyer, and V.~Stoyanov, ``Unsupervised
  cross-lingual representation learning at scale,'' \emph{arXiv preprint
  arXiv:1911.02116}, 2019.

\bibitem{davis2011inferring}
C.~A. Davis~Jr, G.~L. Pappa, D.~R.~R. De~Oliveira, and F.~de~L.~Arcanjo,
  ``Inferring the location of twitter messages based on user relationships,''
  \emph{Transactions in GIS}, vol.~15, no.~6, pp. 735--751, 2011.

\bibitem{devlin2018bert}
J.~Devlin, M.-W. Chang, K.~Lee, and K.~Toutanova, ``Bert: Pre-training of deep
  bidirectional transformers for language understanding,'' \emph{arXiv preprint
  arXiv:1810.04805}, 2018.

\bibitem{dutt2018savitr}
R.~Dutt, K.~Hiware, A.~Ghosh, and R.~Bhaskaran, ``Savitr: A system for
  real-time location extraction from microblogs during emergencies,'' in
  \emph{Companion Proceedings of the The Web Conference 2018}, 2018, pp.
  1643--1649.

\bibitem{gelernter2011geo}
J.~Gelernter and N.~Mushegian, ``Geo-parsing messages from microtext,''
  \emph{Transactions in GIS}, vol.~15, no.~6, pp. 753--773, 2011.

\bibitem{ikawa2012location}
Y.~Ikawa, M.~Enoki, and M.~Tatsubori, ``Location inference using microblog
  messages,'' in \emph{Proceedings of the 21st international conference on
  world wide web}, 2012, pp. 687--690.

\bibitem{imran2022tbcov}
M.~Imran, U.~Qazi, and F.~Ofli, ``Tbcov: Two billion multilingual covid-19
  tweets with sentiment, entity, geo, and gender labels,'' \emph{Data}, vol.~7,
  no.~1, p.~8, 2022.

\bibitem{kinsella2011m}
S.~Kinsella, V.~Murdock, and N.~O'Hare, ``" i'm eating a sandwich in glasgow"
  modeling locations with tweets,'' in \emph{Proceedings of the 3rd
  international workshop on Search and mining user-generated contents}, 2011,
  pp. 61--68.

\bibitem{fpsb-jz61-20}
\BIBentryALTinterwordspacing
R.~Lamsal, ``Coronavirus (covid-19) geo-tagged tweets dataset,'' 2020.
  [Online]. Available: \url{https://dx.doi.org/10.21227/fpsb-jz61}
\BIBentrySTDinterwordspacing

\bibitem{781w-ef42-20}
\BIBentryALTinterwordspacing
------, ``Coronavirus (covid-19) tweets dataset,'' 2020. [Online]. Available:
  \url{https://dx.doi.org/10.21227/781w-ef42}
\BIBentrySTDinterwordspacing

\bibitem{lamsal2021design}
------, ``Design and analysis of a large-scale covid-19 tweets dataset,''
  \emph{Applied Intelligence}, vol.~51, no.~5, pp. 2790--2804, 2021.

\bibitem{lamsalabproblem}
R.~Lamsal, A.~Harwood, and M.~R. Read, ``Addressing the location a/b problem on
  twitter: The next generation location inference research,'' in
  \emph{Proceedings of the 6th ACM SIGSPATIAL LocalRec}, 2022, pp. 1--4.

\bibitem{lamsal2022socially}
------, ``Socially enhanced situation awareness from microblogs using
  artificial intelligence: A survey,'' \emph{ACM Computing Surveys (CSUR)},
  2022.

\bibitem{lamsal2022twitter}
------, ``Twitter conversations predict the daily confirmed covid-19 cases,''
  \emph{Applied Soft Computing}, vol. 129, p. 109603, 2022.

\bibitem{lau2017end}
J.~H. Lau, L.~Chi, K.-N. Tran, and T.~Cohn, ``End-to-end network for twitter
  geolocation prediction and hashing,'' \emph{arXiv preprint arXiv:1710.04802},
  2017.

\bibitem{li2014fine}
C.~Li and A.~Sun, ``Fine-grained location extraction from tweets with temporal
  awareness,'' in \emph{Proceedings of the 37th international ACM SIGIR
  conference on Research \& development in information retrieval}, 2014, pp.
  43--52.

\bibitem{li2014effective}
G.~Li, J.~Hu, J.~Feng, and K.-l. Tan, ``Effective location identification from
  microblogs,'' in \emph{2014 IEEE 30th International Conference on Data
  Engineering}.\hskip 1em plus 0.5em minus 0.4em\relax IEEE, 2014, pp.
  880--891.

\bibitem{li2018location}
P.~Li, H.~Lu, N.~Kanhabua, S.~Zhao, and G.~Pan, ``Location inference for
  non-geotagged tweets in user timelines,'' \emph{IEEE Transactions on
  Knowledge and Data Engineering}, vol.~31, no.~6, pp. 1150--1165, 2018.

\bibitem{lingad2013location}
J.~Lingad, S.~Karimi, and J.~Yin, ``Location extraction from disaster-related
  microblogs,'' in \emph{Proceedings of the 22nd international conference on
  world wide web}, 2013, pp. 1017--1020.

\bibitem{liu2019roberta}
Y.~Liu, M.~Ott, N.~Goyal, J.~Du, M.~Joshi, D.~Chen, O.~Levy, M.~Lewis,
  L.~Zettlemoyer, and V.~Stoyanov, ``Roberta: A robustly optimized bert
  pretraining approach,'' \emph{arXiv preprint arXiv:1907.11692}, 2019.

\bibitem{mahmud2012tweet}
J.~Mahmud, J.~Nichols, and C.~Drews, ``Where is this tweet from? inferring home
  locations of twitter users,'' in \emph{Proceedings of the International AAAI
  Conference on Web and Social Media}, vol.~6, no.~1, 2012, pp. 511--514.

\bibitem{malmasi2015location}
S.~Malmasi and M.~Dras, ``Location mention detection in tweets and
  microblogs,'' in \emph{Conference of the Pacific Association for
  Computational Linguistics}.\hskip 1em plus 0.5em minus 0.4em\relax Springer,
  2015, pp. 123--134.

\bibitem{mchugh2012interrater}
M.~L. McHugh, ``Interrater reliability: the kappa statistic,'' \emph{Biochemia
  medica}, vol.~22, no.~3, pp. 276--282, 2012.

\bibitem{middleton2013real}
S.~E. Middleton, L.~Middleton, and S.~Modafferi, ``Real-time crisis mapping of
  natural disasters using social media,'' \emph{IEEE Intelligent Systems},
  vol.~29, no.~2, pp. 9--17, 2013.

\bibitem{nguyen2020bertweet}
D.~Q. Nguyen, T.~Vu, and A.~T. Nguyen, ``Bertweet: A pre-trained language model
  for english tweets,'' \emph{arXiv preprint arXiv:2005.10200}, 2020.

\bibitem{paraskevopoulos2015fine}
P.~Paraskevopoulos and T.~Palpanas, ``Fine-grained geolocalisation of
  non-geotagged tweets,'' in \emph{Proceedings of the 2015 IEEE/ACM
  International Conference on Advances in Social Networks Analysis and Mining
  2015}, 2015, pp. 105--112.

\bibitem{priedhorsky2014inferring}
R.~Priedhorsky, A.~Culotta, and S.~Y. Del~Valle, ``Inferring the origin
  locations of tweets with quantitative confidence,'' in \emph{Proceedings of
  the 17th ACM conference on Computer supported cooperative work \& social
  computing}, 2014, pp. 1523--1536.

\bibitem{qazi2020geocov19}
U.~Qazi, M.~Imran, and F.~Ofli, ``Geocov19: a dataset of hundreds of millions
  of multilingual covid-19 tweets with location information,'' \emph{SIGSPATIAL
  Special}, vol.~12, no.~1, pp. 6--15, 2020.

\bibitem{sang2003introduction}
E.~F. Sang and F.~De~Meulder, ``Introduction to the conll-2003 shared task:
  Language-independent named entity recognition,'' \emph{arXiv preprint
  cs/0306050}, 2003.

\bibitem{sanh2019distilbert}
V.~Sanh, L.~Debut, J.~Chaumond, and T.~Wolf, ``Distilbert, a distilled version
  of bert: smaller, faster, cheaper and lighter,'' \emph{arXiv preprint
  arXiv:1910.01108}, 2019.

\bibitem{schulz2013multi}
A.~Schulz, A.~Hadjakos, H.~Paulheim, J.~Nachtwey, and M.~M{\"u}hlh{\"a}user,
  ``A multi-indicator approach for geolocalization of tweets,'' in
  \emph{Proceedings of the International AAAI Conference on Web and Social
  Media}, vol.~7, no.~1, 2013, pp. 573--582.

\bibitem{vaswani2017attention}
A.~Vaswani, N.~Shazeer, N.~Parmar, J.~Uszkoreit, L.~Jones, A.~N. Gomez,
  {\L}.~Kaiser, and I.~Polosukhin, ``Attention is all you need,''
  \emph{Advances in neural information processing systems}, vol.~30, 2017.

\bibitem{wolf2019huggingface}
T.~Wolf, L.~Debut, V.~Sanh, J.~Chaumond, C.~Delangue, A.~Moi, P.~Cistac,
  T.~Rault, R.~Louf, M.~Funtowicz \emph{et~al.}, ``Huggingface's transformers:
  State-of-the-art natural language processing,'' \emph{arXiv preprint
  arXiv:1910.03771}, 2019.

\bibitem{yang2019xlnet}
Z.~Yang, Z.~Dai, Y.~Yang, J.~Carbonell, R.~R. Salakhutdinov, and Q.~V. Le,
  ``Xlnet: Generalized autoregressive pretraining for language understanding,''
  \emph{Advances in neural information processing systems}, vol.~32, 2019.

\bibitem{yuan2013and}
Q.~Yuan, G.~Cong, Z.~Ma, A.~Sun, and N.~M. Thalmann, ``Who, where, when and
  what: discover spatio-temporal topics for twitter users,'' in
  \emph{Proceedings of the 19th ACM SIGKDD international conference on
  Knowledge discovery and data mining}, 2013, pp. 605--613.

\bibitem{zubiaga2017towards}
A.~Zubiaga, A.~Voss, R.~Procter, M.~Liakata, B.~Wang, and A.~Tsakalidis,
  ``Towards real-time, country-level location classification of worldwide
  tweets,'' \emph{IEEE Transactions on Knowledge and Data Engineering},
  vol.~29, no.~9, pp. 2053--2066, 2017.

\end{thebibliography}

\newpage

\appendix
\section{Appendix}

\begin{footnotesize}
\label{app-incorrect-classifications}
Below are some examples from \textit{tweets origin location evidence dataset} which were incorrectly classified by \textit{locBERT}. Labels: $0$ represents \textit{true origin} class, and $1$ represents \textit{low evidence} class.

\textbf{tweet}: 15 more deaths confirmed here, Italy to re-open in June: Today's Covid-19 main points: Here are the main points to know about Covid-19 in Ireland and around the world today. HTTPURL, \textbf{label}: 0

\textbf{tweet}: I cried today as I am missing my Amma(mum)and home for the very first time since I moved away some 13 years to my new home in England. I think this lockdown really brought me close to my amma, I really enjoyed the… HTTPURL, \textbf{label}: 0

\textbf{tweet}: Thankful for biancabeltran and @USER today! They came down to the Historic 18th \& Vine Jazz District, Kansas City to speak to me about COVID-19 and how it's effecting my business and the Industry as a whole but… HTTPURL, \textbf{label}: 0

\textbf{tweet}: I was on AIT Kakaaki this morning and shared further thoughts on the Covid-19 Control efforts in Nigeria. Some points I made were as follows; 1. Catholic Church donating their hospitals is a good gesture but they… HTTPURL, \textbf{label}: 0

\textbf{tweet}: In fact, today I completed the Sweat Pocari competition in Bandung, canceled because of the Corona virus that shook the world. And finally I can reach 21 km of the city of Medan. \#longrun… HTTPURL, \textbf{label}: 0

\textbf{tweet}: Taharka. ICU doctor I met at the protest last night in Brooklyn. We spoke about George Floyd, about everyday racism in America and about Covid-19. \#georgefloyd \#brooklynprotests \#blacklivesmatter \#blm \#journalism… HTTPURL, \textbf{label}: 0

\textbf{tweet}: An early morning hike at Ramapo. It was great to see how seriously people are taking social distancing and mask wearing. Every time I passed people we would put on masks. 95\% of time your… HTTPURL, \textbf{label}: 1

\textbf{tweet}: They’re hosting a \#COVID19 vaccine clinic today.. (at @USER in Berkeley, CA) HTTPURL, \textbf{label}: 1

\textbf{tweet}: Getting ready for my virtual happy hour, with friends and family in Manhattan, Brooklyn and Jersey City!\#covid19 \#corona \#virtualhappyhour \#virtual5a7 \#nyccrowd \#lockdown2019 \#denmark \#amager \#kimcrawford… HTTPURL, \textbf{label}: 1

\textbf{tweet}: Palo Alto showed up to rally and march in celebration of Juneteenth today while being mindful and social distancing. The energy and determination was palpable. My fellow white people, we are responsible for systemic… HTTPURL, \textbf{label}: 1

\textbf{tweet}: FREE COVID-19 mobile screening and testing TODAY! Monday August 24, 2020 Iglesias de Camino 1801 Walkup Avenue, Monroe, NC 28112 8 am - 2 pm This event is for anyone who is symptomatic… HTTPURL, \textbf{label}: 1

\textbf{tweet}: Happy May 1!! Today I would have began walking the Camino de Santiago. It is a 500 mile walk from the French side of the Pyrenees, across northern Spain, ending in Santiago de Compestella. Since Covid-19 cancelled… HTTPURL, \textbf{label}: 1

\end{footnotesize}

\end{document}